\pdfoutput=1
\documentclass[11pt,a4paper]{article}
\usepackage[hyperref]{naaclhlt2019}
\usepackage{times}
\usepackage{latexsym}

\usepackage{url}

\aclfinalcopy 
\def\aclpaperid{1806} 

\usepackage{comment}
\usepackage{soul}

\usepackage{helvet}  
\usepackage{courier}  
\usepackage{url}  
\usepackage{graphicx}  
\usepackage{booktabs}
\usepackage{tabularx}
\usepackage{comment}

\usepackage{microtype}
\usepackage{graphicx}
\usepackage{subfigure}
\usepackage{booktabs} 
\usepackage{wrapfig}

\usepackage[utf8]{inputenc} 
\usepackage[T1]{fontenc}    
\usepackage{url}            
\usepackage{amsfonts}       
\usepackage{nicefrac}       
\graphicspath{{figures/}}
\usepackage{xcolor}
\usepackage{adjustbox}
\usepackage{bm}
\usepackage{amsmath}
\usepackage{amssymb}
\usepackage{algorithm}
\usepackage{algorithmic}
\usepackage{multirow}
\usepackage{adjustbox}

\usepackage{bm}

\newcommand{\RN}[1]{%
	\textup{\lowercase\expandafter{\it \romannumeral#1}}%
}

\usepackage{algorithm}
\usepackage{algorithmic}

\usepackage{amsmath}
\usepackage{amsfonts}
\usepackage{amssymb}

\newcommand{\beq}{\vspace{0mm}\begin{equation}}
\newcommand{\eeq}{\vspace{0mm}\end{equation}}
\newcommand{\beqs}{\vspace{0mm}\begin{eqnarray}}
\newcommand{\eeqs}{\vspace{0mm}\end{eqnarray}}
\newcommand{\barr}{\begin{array}}
\newcommand{\earr}{\end{array}}

\newcommand{\cv}{{\boldsymbol c}}
\newcommand{\dv}{{\boldsymbol d}}

\newcommand{\fv}{{\boldsymbol f}}

\newcommand{\lv}{{\boldsymbol l}}

\newcommand{\rv}{{\boldsymbol r}}
\newcommand{\sv}{{\boldsymbol s}}

\newcommand{\vv}{{\boldsymbol v}}
\newcommand{\wv}{{\boldsymbol w}}
\newcommand{\Wv}{{\boldsymbol W}}

\newcommand{\Uv}{{\boldsymbol U}}
\newcommand{\xv}{{\boldsymbol x}}

\usepackage[bottom]{footmisc}

 \def\argmax{\mathop{\rm arg\,max}}

\title{Context-Dependent Semantic Parsing over  Temporally Structured Data}

\author{Charles Chen and Razvan Bunescu\\
  School of Electrical Engineering and Computer Science, Ohio University \\
  {\tt lc971015@ohio.edu, bunescu@ohio.edu}}

\date{}

\begin{document}
\maketitle

\begin{abstract}
We describe a new semantic parsing setting that allows users to query the system using both natural language questions and actions within a graphical user interface. Multiple time series belonging to an entity of interest are stored in a database and the user interacts with the system to obtain a better understanding of the entity's state and behavior, entailing sequences of actions and questions whose answers may depend on previous factual or navigational interactions. We design an LSTM-based encoder-decoder architecture that models context dependency through copying mechanisms and multiple levels of attention over inputs and previous outputs. When trained to predict tokens using supervised learning, the proposed architecture substantially outperforms standard sequence generation baselines. Training the architecture using policy gradient leads to further improvements in performance, reaching a sequence-level accuracy of 88.7\% on artificial data and 74.8\% on real data.
\end{abstract}

\section{Introduction and Motivation}
\label{section:introduction}

Wearable sensors are being increasingly used in medicine to monitor important physiological parameters. Patients with type I diabetes, for example, wear a sensor inserted under the skin which provides measurements of the interstitial blood glucose level (BGL) every 5 minutes. Sensor bands provide a non-invasive solution to measuring additional physiological parameters, such as temperature, skin conductivity, heart rate, and acceleration of body movements. Patients may also self-report information about discrete life events such as meals, sleep, or stressful events, while an insulin pump automatically records two types of insulin interventions: a continuous stream of insulin called the basal rate, and discrete self-administered insulin dosages called boluses. The data acquired from sensors and patients accumulates rapidly and leads to a substantial data overload for the health provider.

\begin{figure}[t]
	\centering
	\includegraphics[width=\columnwidth]{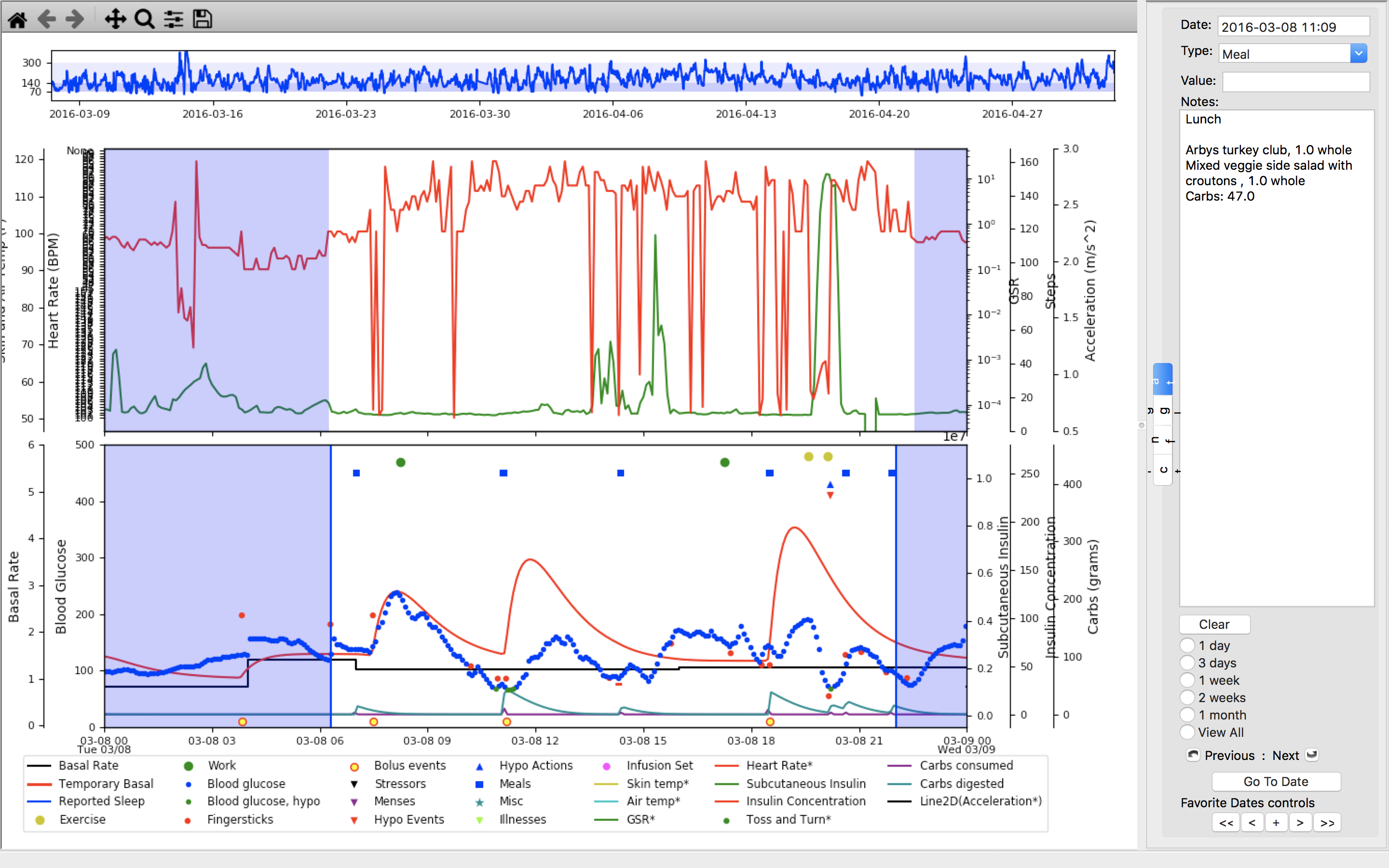}
	\caption {GUI window displaying 1 day worth of data.}
	\label{fig:gui}
\end{figure}

To help doctors more easily browse the wealth of generated patient data, we built a graphical user interface (GUI) that displays the various time series of measurements corresponding to a patient. As shown in Figure~\ref{fig:gui}, the GUI displays the data corresponding to one day, whereas buttons allow the user to move to the next or previous day. While the graphical interface was enthusiastically received by doctors, it soon became apparent that the doctor-GUI interaction could be improved substantially if the tool also allowed for natural language (NL) interactions. Most information needs are highly contextual and local. For example, if the blood glucose spiked after a meal, the doctor would often want to know more details about the meal or about the bolus that preceded the meal. The doctor often found it easier to express their queries in natural language (e.g. ``show me how much he ate'', ``did he bolus before that''), resulting in a sub-optimal situation where the doctor would ask this type of {\it local questions} in English while a member of our team would perform the clicks required to answer the question, e.g. click on the meal event, to show details such as amount of carbohydrates. Furthermore, there were also {\it global questions}, such as ``How often does the patient go low in the morning and the evening'', whose answers would require browsing the entire patient history in the worst case, which would be very inefficient. This motivated us to start work on a new system component that would allow the doctor to interact using both natural language queries and direct actions within the GUI. A successful solution to the task described in this paper has the potential for applications in many areas of medicine where sensor data and life events are pervasive. Intelligent user interfaces for the proposed task will also benefit the exploration and interpretation of data in other domains such as experimental physics, where large amounts of time series data are generated from high-throughput experiments.

\section{Task Definition}
\label{section:definition}

Given an input from the user (a NL query or a direct GUI interaction), the aim is to parse it into a logical form representation that can be run by an inference engine in order to automatically extract the answer from the database. Table~\ref{tab:examples} shows sample inputs paired with their logical forms. For each input, the examples also show relevant previous inputs from the interaction sequence.
\begin{table}[t]
\setlength{\tabcolsep}{0em}
\small{
\centering
\begin{tabular}{l}
\hline
\multicolumn{1}{c}{\bf Example 1} \\
\hline
Click on Exercise event at 9:29am. \\
$Click(e) \wedge e.type = \text{Exercise} \wedge e.time = \text{9:29am}$\\[0.5em]
Click on Miscellaneous event at 9:50am \\
$Click(e) \wedge e.type = \mbox{Misc} \wedge e.time = \mbox{9:50am} $ \\[0.5em]
Q$_1$: What was she doing mid afternoon \\
   \hspace{9em} when her heart rate went up? \\
$Answer(e) \wedge Behavior(e_1.value, \mbox{Up})$ \\ 
  \hspace{5em} $\wedge Around(e.time, e_1.time)$ \\
  \hspace{5em} $\wedge e.type == DiscreteType$ \\
  \hspace{5em} $\wedge e1.type == HeartRate$ \\
  \hspace{5em} $\wedge e1.time == MidAfternoon()$ \\[0.5em]
Q$_2$: What time did that start? \\
$Answer(e(-1).time)$ \\
\hline
\multicolumn{1}{c}{\bf Example 2} \\
\hline
Click on Bolus at 8:03pm. \\
$Click(e) \wedge e.type = \mbox{Bolus} \wedge e.time = \mbox{8:03pm}$ \\[0.5em]
Q$_3$: What did she eat for her snack? \\
$Answer(e.food) \wedge e.kind == Snack$ \\
\hline
\multicolumn{1}{c}{\bf Example 3} \\
\hline
Click on Exercise at 7:52pm. \\
$Click(e) \wedge e.type = \mbox{Exercise} \wedge e.time = \mbox{7:52pm}$ \\[0.5em]
Q$_4$: What did she do then? \\
$Answer(e(-1).kind)$\\[0.5em]
Q$_5$: Did she take a bolus before then? \\
$Answer(Any(d.type == Bolus$\\
\hspace{2em} $\wedge Before(d.time, e(-1).time)))$\\ 
\hline
\multicolumn{1}{c}{\bf Example 4} \\
\hline
Q$_6$: What is the first day they have heart rate reported?\\
$Answer(e.date)$\\
\hspace{2em} $\wedge Order(e, 1, Sequence(d, d.type == HeartRate))$\\
\hline
\multicolumn{1}{c}{\bf Example 5} \\
\hline
Q$_7$: Is there another day he goes low in the morning?\\
$Answer(Any(Hypo(d1) \wedge x! = CurrentDate$\\
\hspace{2em}$\wedge x.type == Date \wedge d1.time == Morning(x))$\\
\hline    
\end{tabular}
}
\caption{Examples of interactions and logical forms.}
\label{tab:examples}
\end{table}
In the following sections we describe a number of major features that, on their own or through their combination, distinguish this task from  other semantic parsing tasks.

\subsection{Time is essential}

All events and measurements in the knowledge base are organized in time series. Consequently, many queries contain time expressions, such as the relative ``midnight'' or the coreferential ``then'', and temporal relations between relevant entities, expressed through words such as ``after'' or ``when''. This makes processing of temporal relations essential for a good performance. Furthermore, the GUI serves to anchor the system in time, as most of the information needs expressed in local questions are relative to the day shown in the GUI, or the last event that was clicked.

\subsection{GUI interactions vs. NL questions}

The user can interact with the system 1) directly within the GUI (e.g. mouse clicks); 2) through natural language questions; or 3) through a combination of both, as shown in Examples 1 and 2 in Table~\ref{tab:examples}. Although the result of every direct interaction with the GUI can also be obtained using natural language questions, sometimes it can be more convenient to use the GUI directly, especially when all events of interest are in the same area of the screen and thus easy to move the mouse or hand from one to the other. For example, a doctor interested in what the patient ate that day can simply click on the blue squares at the top of the bottom pane in Figure~\ref{fig:gui}, one after another. Sometimes a click can be used to anchor the system at a particular time during the day, after which the doctor can ask short questions implicitly focused on that region in time. An example of such hybrid behavior is shown in Example 2, where a click on a Bolus event is followed by a question about a snack, which implicitly should be the meal right after the bolus.

\subsection{Factual queries vs. GUI commands}

Most of the time, doctors have information needs that can be satisfied by clicking on an event shown in the GUI or by asking factual questions about a particular event of interest from that day. In contrast, a different kind of interaction happens when the doctor wants to change what is shown in the tool, such as toggling on/off particular time series (e.g. ``toggle on heart rate''), or navigating to a different day (e.g. ``go to next day'', ``look at the previous day''). Sometimes, a question may be a combination of both, as in ``What is the first day they have a meal without a bolus?'', for which the expectation is that the system navigates to that day and also clicks on the meal event to show additional information and anchor the system at the time of that meal.

\subsection{Sequential dependencies}

The user interacts with the system through a sequence of questions or clicks. The logical form of a question, and implicitly its answer, may depend on the previous interaction with the system. Examples 1 to 3 in Table~\ref{tab:examples} are all of this kind. In example 1, the pronoun ``that'' in question 2 refers to the answer to question 1. In example 2, the snack refers to the meal around the time of the bolus event that was clicked previously -- this is important, as there may be multiple snacks that day. In example 3, the adverb ``then'' in question 5  refers to the time of the event that is the answer of the previous question. As can be seen from these examples, sequential dependencies can be expressed as coreference between events from different questions. Coreference may also happen within questions, as in question 4 for example. Overall, solving coreferential relations will be essential for good performance.

\section{Semantic Parsing Datasets}
\label{section:dataset}

To train and evaluate semantic parsing approaches, we created two datasets of sequential interactions: a dataset of real interactions (Section~\ref{sec:real}) and a much larger dataset of artificial interactions (Section~\ref{sec:artificial}).

\subsection{Real Interactions}
\label{sec:real}

We recorded interactions with the GUI in real time, using data from 9 patients, each with around 8 weeks worth of time series data. In each recording session, the  tool was loaded with data from one patient and the physician was instructed to explore the data in order to understand the patient behavior as usual, by asking NL questions or interacting directly with the GUI. Whenever a question was asked, a member of our study team found the answer by navigating in and clicking on the corresponding event. After each session, the question segments were extracted manually from the speech recordings, transcribed, and timestamped. All direct interactions (e.g. mouse clicks) were recorded automatically by the tool, timestamped, and exported into an XML file. The sorted list of questions and the sorted list of mouse clicks were then merged using the timestamps as key, resulting in a chronologically sorted list of questions and GUI interactions. Mouse clicks were automatically translated into logical forms, whereas questions were parsed into logical forms manually.


\begin{table}[t]
\setlength{\tabcolsep}{0em}
\small{
\centering
\begin{tabular}{l}
\hline
\multicolumn{1}{c}{\bf Event Types}\\
\hline
\textit{Physiological Parameters}: \\
\hspace{0.5em} BGL, BasalRate, TemporaryBasal, Carbs, GSR, InfusionSet, \\
\hspace{0.5em} AirTemperature, SkinTemperature, HeartRate, StepCount.\\
\textit{Life Events}:\\
\hspace{0.5em} FingerSticks, Bolus, Hypo, HypoAction, Misc, Illness,\\
\hspace{0.5em} Meal, Exercise, ReportedSleep, Wakeup, Work, Stressors.\\
\hline
\multicolumn{1}{c}{\bf Constants}\\
\hline
Up, Down, On, Off, Monday, Tuesday, ..., Sunday.\\
\hline
\multicolumn{1}{c}{\bf Functions}\\
\hline
$Interval(t_1, t_2)$, $Before(t)$, $After(t)$, ... \\
\hspace{0.5em} return corresponding intervals (default lengths). \\
$Morning([d])$, $Afternoon([d])$, $Evening([d])$, ... \\
\hspace{0.5em} return corresponding intervals for day $d$. \\
$WeekDay(d)$:\\
\hspace{0.5em} return the day of the week of date $d$. \\
$Sequence(var, statements)$: \\
\hspace{0.5em} return a chronologically ordered sequence \\
\hspace{0.5em} of possible values for $var$ that satisfy $statements$.\\
$Count(var[,statements])$:\\
\hspace{0.5em} returns the number of possible values \\
\hspace{0.5em} for $var$ that satisfy $statements$.\\
\hline
\multicolumn{1}{c}{\bf Predicates}\\
\hline
$Answer(e)$, $Click(e)$ \\ 
${Morning(t)}$, ${Afternoon(t)}$, $Evening(t)$,  ... \\
${Overlap(t_1, t_2)}$, ${Before(t_1, t_2)}$, ${Around(t_1, t_2)}$, ... \\

${Behavior(variable,direction)}$: \\
\hspace{0.5em} whether $variable$ increases, if $direction$ is $Up$,\\ 
\hspace{0.5em} (or decrease if $direction$ is $Down$).\\

$High(variable)$, $Low(variable)$: \\
\hspace{0.5em} whether $variable$ has some low value. \\

${Order(event,ordinal,sequence[,attribute])}$:\\
\hspace{0,5em} whether the $event$ is at place $ordinal$ in $sequence$\\
\hline
\multicolumn{1}{c}{\bf Commands}\\
\hline
DoClick, DoToggle, DoSetDate, DoSetTime, ...\\
\hline
\end{tabular}
}
\caption{Vocabulary for logical forms.}
\label{tab:vocabulary}
\end{table}

A snapshot of the vocabulary for logical forms is shown in Table~\ref{tab:vocabulary}, showing the Event Types, Constants, Functions, Predicates, and Commands. Every life event or physiological measurement stored in the database is represented in the logical forms as an event object $e$ with 3 major attributes: $e.type$, $e.date$, and $e.time$. Depending on its type, an event object may contain additional fields. For example, if $e.type = BGL$, then it has an attribute $e.value$. If $e.type = Meal$, then it has attributes $e.food$ and $e.carbs$. We use $e(-i)$ to represent the event appearing in the $i^{th}$ previous logical form (LF). Thus, to reference the event mentioned in the previous LF, we use $e(-1)$, as shown for question Q$_5$. If more than one event appears in the previous LF, we use an additional index $j$ to match the event index in the previous LF. Coreference between events is represented simply using the equality operator, e.g. $e = e(-1)$.
The dataset contains logical forms for 237 interactions: 74 mouse clicks and 163 NL queries.

\subsection{Artificial Interactions}
\label{sec:artificial}

The number of annotated real interactions is too small for training an effective semantic parsing model. To increase the number of training examples, we designed and implemented an artificial data generator that simulates user-GUI interactions, with sentence \textit{templates} defining the skeleton of each entry in order to maintain high-quality sentence structure and grammar. This approach is similar to \cite{weston:babi15}, with the difference that we need a much higher degree of variation such that the machine learning model does not memorize all possible sentences, and consequently a much richer template database. We therefore implemented a \textit{template language} with recursive grammar, that can be used to define as many templates and generate as many data examples as desired. We used the same vocabulary as for the real interactions dataset. To generate contextual dependencies (e.g. event coreference), the implementation allows for more complex \textit{combo templates} where a sequence of templates are instantiated together. A more detailed description of the template language and the simulator implementation is given in ~\cite{chen2019physician} and Appendix A, together with illustrative examples. The simulator was used to generate 1,000 interactions and their logical forms: 312 mouse clicks and 688 NL queries.

\section{Baseline Models for Semantic Parsing}
\label{section:baseline}

This section describes two baseline models: a standard LSTM encoder-decoder for sequence generation {\it SeqGen} (Section~\ref{sec:seqgen}) and its attention-augmented version {\it SeqGen+Att2In} (Section~\ref{sec:att2in}). This last model will be used later in Section~\ref{sec:architecture} as a component in the context-dependent semantic parsing architecture.

\subsection{\textit{SeqGen}}
\label{sec:seqgen}

\begin{figure}[htbp]
{\includegraphics[width=\columnwidth]{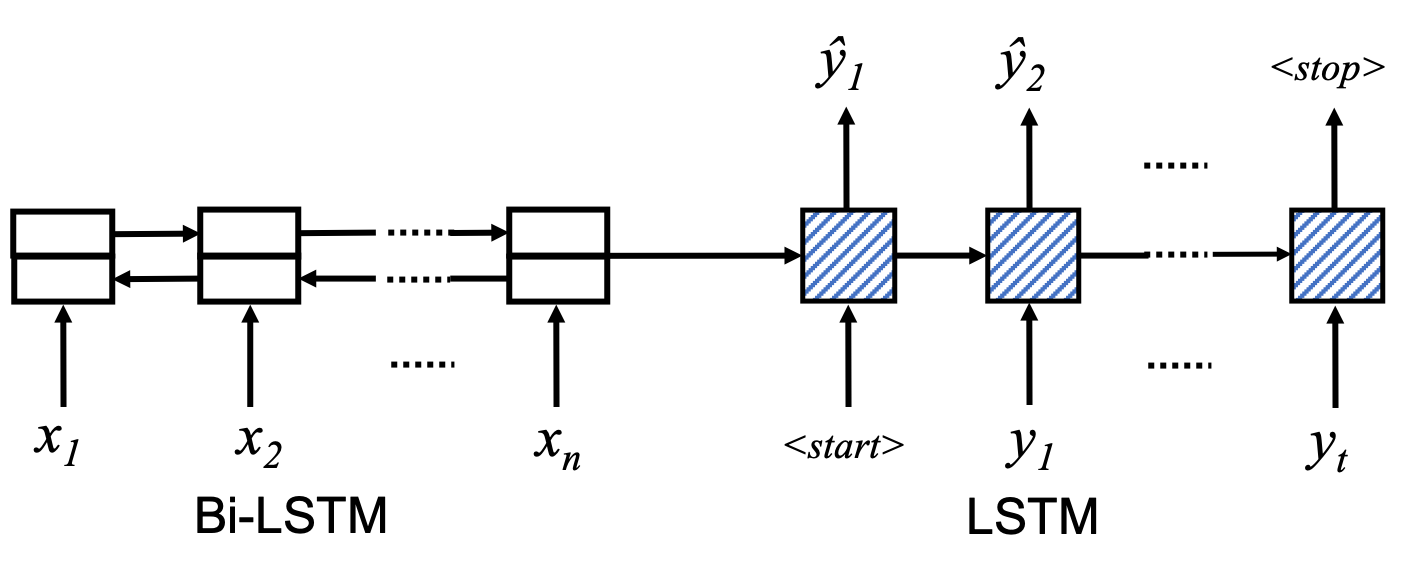}}
\caption{The \textit{SeqGen} model takes a sequence of interactions as input $X =x_1, \ldots, x_n$ and encodes it with a Bi-LSTM (left). The decoder LSTM (right) generates a logical form $\hat{Y} = \hat{y}_1, \ldots, \hat{y}_T$.}
\vspace{-4mm}
\label{fig:seq2seq}
\end{figure}

As shown in Figure~\ref{fig:seq2seq}, the sequence-generation model \textit{SeqGen} uses Long Short-Term Memory (LSTM)~\cite{hochreiter1997long} units in an encoder-decoder architecture~\cite{bahdanau2016actor,cho2014learning}, composed of a bi-directional LSTM for the encoder over the input sequence $X$ and an LSTM for the decoder of the output LF sequence $Y$. We use $Y_t = y_1, \ldots, y_t$ to denote the sequence of output tokens up to position $t$. We use $\hat{Y}$ to denote the generated logical form.

The initial state $\sv_0$ is created by running the bi-LSTM encoder over the input sequence $X$ and concatenating the last hidden states. Starting from the initial hidden state $\sv_0$, the decoder produces a sequence of states $\sv_1,\ldots,\sv_T$, using embeddings  $e(y_t)$ to represent the previous tokens in the sequence.  A softmax is used to compute token probabilities at each position as follows:
\begin{align}
p(y_t|Y_{t-1}, X) & = softmax(\Wv_h \sv_{t}) \\
\sv_t & = h(\sv_{t - 1}, e(y_{t-1})) \nonumber
\end{align}    
The transition function $h$ is implemented by the LSTM unit.


\subsection{\textit{SeqGen+Att2In}}
\label{sec:att2in}

\begin{figure}[thp]
{\includegraphics[width=\columnwidth]{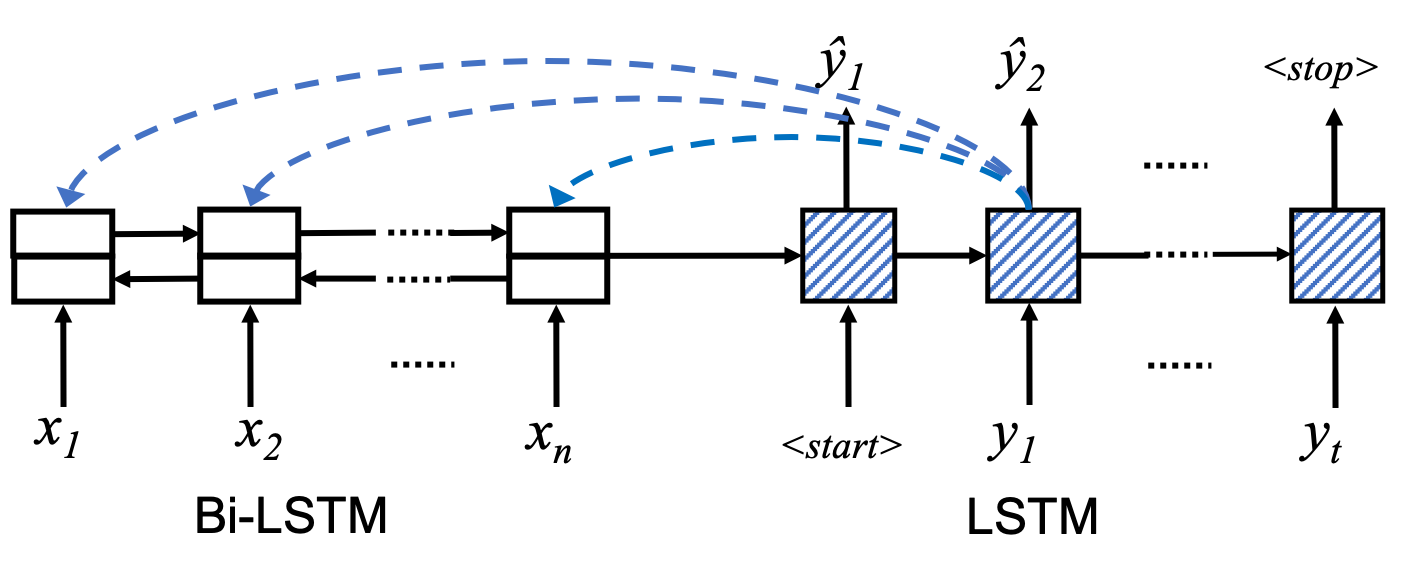}}
\caption{The \textit{SeqGen+Att2In} model augments the \textit{SeqGen} model with an attention mechanism. At each decoding step $t$, it attends to all input tokens in order to compute a context vector $\dv_t$.}
\vspace{-4mm}
\label{fig:1att}
\end{figure}

This model (Figure~\ref{fig:1att}) is similar to \textit{SeqGen}, except that it attends to the current input (NL query or mouse click) during decoding. Equation~\ref{eq:attention1} defines the corresponding attention mechanism  \textit{Att2In} used to create the context vector $\dv_t$:
\begin{align}
e_{tj} & = \vv_{a}^T \tanh(\Wv_{a} \fv_{j} + \Uv_{a} \sv_{t-1}) \label{eq:attention1}  \\
\alpha_{tj} & = \frac{\exp(e_{tj})}{\sum_{k=1}^{m}\exp(e_{tk})} \nonumber,   
   \:\:\:\:\:\dv_t\!=\!\cv_t\!=\!\sum_{j=1}^{n}\alpha_{tj}\fv_j \nonumber 
\end{align}
Here $\fv_j$ is the $j$-th hidden states for Bi-LSTM corresponding to $\xv_j$ and $\alpha_{tj}$ is an attention weight. 
Both the context vector $\dv_t$ and $\sv_t$ are used to predict the next token $\hat{y}_t$ in the logical form:
\begin{align*}
\hat{y}_t & \sim \text{softmax}(\Wv_h \sv_t + \Wv_d \dv_t)
\end{align*}

\section{Context-Dependent Semantic Parsing}
\label{sec:architecture}

\begin{figure*}[htbp]
\centerline{\includegraphics[width=\textwidth]{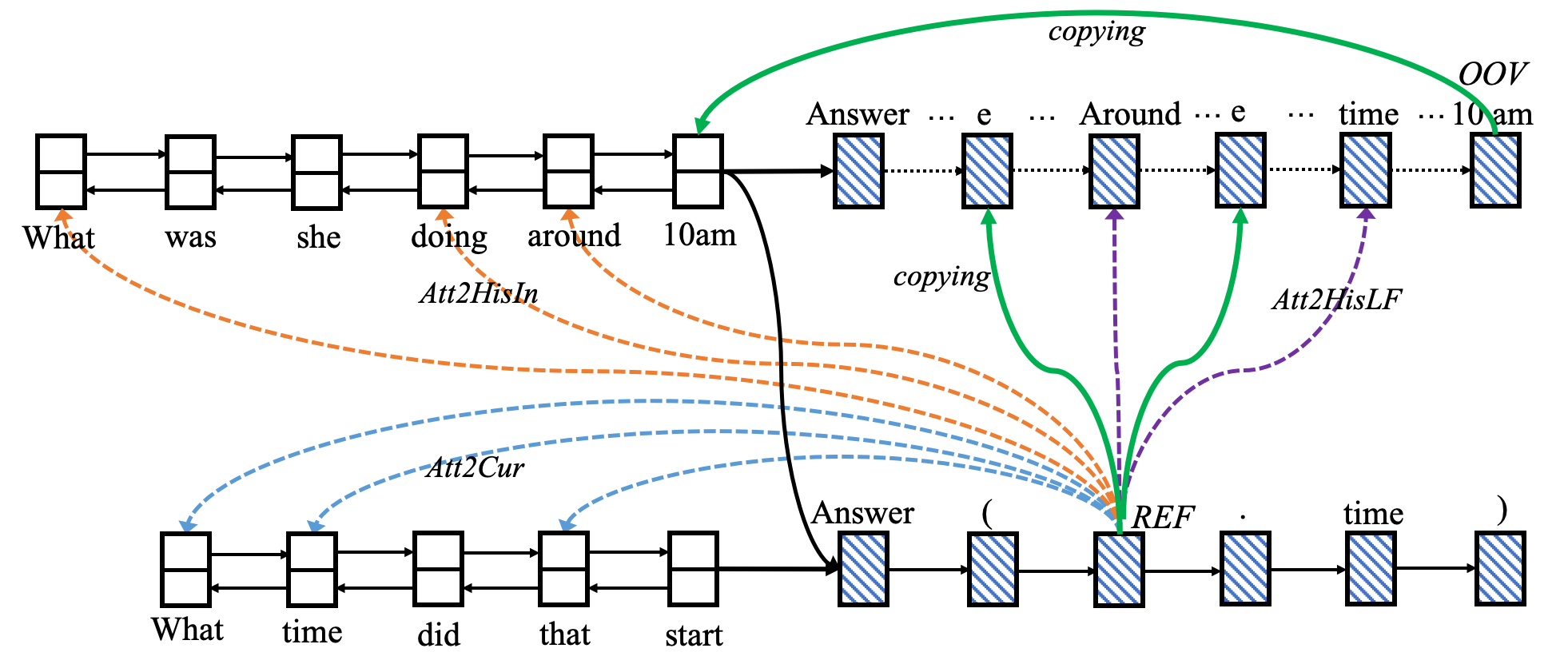}}
\caption{Context-dependent semantic parsing architecture. We use a Bi-LSTM (left) to encode the input and a LSTM (right) as the decoder. 
We show only parts of the LF to save space. The complete generated LF at time T-1 is $Y^{-1}$ = [\textit{Answer, (, e, ), $\wedge$, Around, (, e, ., time, \textit{OOV}, ), $\wedge$, e, ., type, ==, DiscreteType}]. The token {\it 10am} is copied from the input to replace the generated \textit{OOV} token (solid green arrow). The complete generated LF at time T is $Y$ = [\textit{Answer, (, \textit{REF}, ., time, )}]. The entity token {\it e} is copied from the previous LF to replace the generated \textit{REF} token (solid green arrow). Orange dash arrows attend to historical input. Blue dash arrows attend to current input. Purple dash arrows attend to previous logical form.}
\vspace{-4mm}
\label{fig:3att}
\end{figure*}

In Figure~\ref{fig:3att} we show our proposed semantic parsing model,  \textit{SP+Att2All+Copy} (\textit{SPAAC}). Similar to the baseline models, we use a bi-directional LSTM to encode the input and another LSTM as the decoder. Context-dependency is modeled using two types of mechanisms: {\it attention} and {\it copying}. The attention mechanism (Section~\ref{sec:attention}) is comprised of 3 models: \textit{Att2HisIn} attending to the previous input, \textit{Att2HisLF} attending to the previous logical form, and the \textit{Att2In} introduced in Section~\ref{sec:att2in} that attends to the current input. The copying mechanism (Section~\ref{sec:copy}) is comprised of two models: one for handling unseen tokens, and one for handling coreference to events in the current and previous logical forms.

\subsection{Attention Mechanisms}
\label{sec:attention}

At decoding step $t$, the \textit{Att2HisIn} attention model computes the context vector $\hat{\cv_t}$ as follows:
\begin{align}
\hat{e}_{tk} & = \vv_b^T \tanh(\Wv_b \rv_{k} + \Uv_b \sv_{t-1}) \label{eq:attention2} \\
\beta_{tk} & = \frac{\exp(\hat{e}_{tk})}{\sum_{l=1}^{m^2}\exp(\hat{e}_{tl})} \nonumber,
\:\:\:\:\:\hat{\cv_t}  = \sum_{k=1}^{n}\beta_{tk} \cdot \rv_k \nonumber 
\end{align}
where $\rv_k$ is the encoder hidden state corresponding to $\xv_k$ in the previous input $X^{-1}$, $\hat{\cv_t}$ is the context vector, and $\beta_{tk}$ is an attention weight. 

Similarly, the \textit{Att2HisLF} model computes the context vector $\tilde{\cv_t}$ as follows:
\begin{align}
\tilde{e}_{tj} & = {\vv_c}^T \tanh(\Wv_c \lv_{j} + \Uv_c \sv_{t-1}) \label{eq:attention3}  \\
\gamma_{tj} & = \frac{\exp(\tilde{e}_{tj})}{\sum_{j=1}^{n}\exp(\tilde{e}_{tj})} \nonumber,
\:\:\:\:\:\tilde{\cv_t}  = \sum_{j=1}^{n}\gamma_{tj} \cdot \lv_j \nonumber 
\end{align}
where  $\lv_j$ is the $j$-th hidden state of the decoder for the previous logical form $Y^{-1}$.

The context vector used in the decoder is comprised of the context vectors from the three attention models \textit{Att2In}, \textit{Att2HisIn} and \textit{Att2HisLF}:
\begin{align}
\dv_t = concat(\cv_t, \hat{\cv_t}, \tilde{\cv_t})
\end{align}

\subsection{Copying Mechanisms}
\label{sec:copy}

In order to handle out-of-vocabulary (OOV) tokens and coreference (REF) between entities in the current and the previous logical forms, we add two special tokens \textit{OOV} and \textit{REF} to the vocabulary. Inspired by the copying mechanism in~\cite{gu2016incorporating}, we train the model to learn which token in the current input $X=\{x_j\}$ is an OOV by minimizing the following loss:
\begin{align}
L_{oov}(Y) & = -\sum_{t=1}^{Y.l} \sum_{j=1}^{X.l} \log p_o(O_{j} | \sv^{X}_{j}, \sv^{Y}_{t})
\label{eq:copy-loss}
\end{align}
where $X.l$ is the length of current input, $Y.l$ is the length of the current logical form, $\sv^{X}_{j}$ is the LSTM state for $x_j$ and $\sv^{Y}_{t}$ is the LSTM state for $y_t$, $O_j \in \{0, 1\}$ is a label indicating whether $x_{j}$ is an OOV. We use logistic regression to compute the OOV probability, i.e. $p_o(O_{j} = 1 | \sv^{X}_{j}, \sv^{Y}_{t}) = \sigma(\wv^{T}_{o}[\sv^{X}_{j},\sv^{Y}_{t}])$.

Similarly, to solve coreference, the model is trained to learn which entity in the previously generated logical form $\hat{Y}^{-1}=\{\hat{y}_j\}$ is coreferent with the entity in the current logical form by minimizing the following loss:
\begin{align}
L_{ref}(Y) &\! =\! -\sum_{t=1}^{Y.l}\sum_{j=1}^{\hat{Y}^{-1}.l} \log p_r(R_{j} | \sv^{\hat{Y}^{-1}}_{j}\!\!, \sv^{Y}_{t})
\label{eq:coref-loss}
\end{align}
where $\hat{Y}^{-1}.l$ is the length of the previous generated logical form, $Y.l$ is the length of the current logical form, $\sv^{\hat{Y}^{-1}}_{j}$ is the LSTM state at position $j$ in $\hat{Y}^{-1}$ and $\sv^{Y}_{t}$ is the LSTM state for position $t$ in $Y$, and $R_{j} \in \{0, 1\}$ is a label indicating whether $\hat{y}_{j}$ is an entity referred by $y_t$ in the next logical form $Y$. We use logistic regression to compute the coreference probability, i.e. $p_r(R_{j} = 1 | \sv^{\hat{Y}^{-1}}_{j}, \sv^{Y}_{t}) = \sigma(\wv^{T}_{r}[\sv^{\hat{Y}^{-1}}_{j}, \sv^{Y}_{t}])$.

Finally, we use ``Teacher forcing''~\cite{williams1989learning} to train the model to learn which token in the vocabulary (including special tokens \textit{OOV} and \textit{REF}) should be generated, by minimizing the following token generation loss:
\begin{align}
L_{gen}(Y) = -\sum_{t=1}^{{Y}.l} \log p(y_t | {Y}_{t-1}, X)
\label{eq:gen-loss}
\end{align}
where ${Y}.l$ is the length of the current logical form.

\subsection{Supervised Learning: \textit{SPAAC-MLE}}
\label{sec:mle}

The supervised learning model \textit{SPAAC-MLE} is obtained by training the semantic parsing architecture from Figure~\ref{fig:3att} to minimize the sum of the 3 negative log-likelihood losses:
\begin{align}
L_{MLE}(Y)\!=\! L_{gen}(Y)\! +\! L_{oov}(Y)\! + \!L_{ref}(Y)
\label{eq:mle-loss}
\end{align}
At inference time, beam search is used to generate the LF sequence~\cite{ranzato2015sequence, wiseman2016sequence}. During inference, if the generated token at position $t$ is \textit{OOV}, we copy the token from the current input $X$ that has the maximum OOV probability, i.e. $\argmax_j p_o(O_j = 1 | \sv^{X}_{j}, \sv^{Y}_{t})$. Similarly, if the generated entity token at position $t$ is \textit{REF}, we copy the entity token from the previous LF $Y^{-1}$ that has the maximum coreference probability, i.e. $\argmax_j p_r(R_j = 1 | \sv^{Y^{-1}}_{j}, \sv^{Y}_t)$. 


\subsection{Reinforcement Learning: \textit{SPAAC-RL}}
\label{sec:rl}

All models described in this paper are evaluated using sequence-level accuracy, a discrete metric where a generated logical form is considered to be correct if it is equivalent with the ground truth logical form. This is a strict evaluation measure in the sense that it is sufficient for a token to be wrong to invalidate the entire sequence. At the same time, there can be many generated sequences that are correct, e.g. any reordering of the clauses from the ground truth sequence is correct. The large number of potentially correct generations can lead MLE-trained models to have sub-optimal performance~\cite{paulus2017deep,rennie2017self,zeng2016efficient,norouzi2016reward}. Furthermore, although ``teacher forcing''~\cite{williams1989learning} is widely used for training sequence generation models, it leads to {\it exposure bias} ~\cite{ranzato2015sequence}: the network has knowledge of the ground truth LF tokens up to the current token during training, but not during testing, which can lead to propagation of errors at generation time.

Like~\citet{paulus2017deep}, we address these problems by using policy gradient to train a token generation policy that aims to directly maximize sequence-level accuracy. We use the self-critical policy gradient training algorithm proposed by~\citet{rennie2017self}. We model the sequence generation process as a sequence of actions taken according to a policy, which takes an action (token $\hat{y}_t$) at each step $t$ as a function of the current state (history $\hat{Y}_{t-1}$), according to the probability $p(\hat{y}_t | \hat{Y}_{t-1})$. The algorithm uses this probability to define two policies: a greedy, baseline policy $\pi^b$ that takes the action with the largest probability, i.e. $\pi^b(\hat{Y}_{t-1}) = \argmax_{\hat{y}_t} p(\hat{y}_t | \hat{Y}_{t-1})$; and a sampling policy  $\pi^s$ that samples the action according to the same distribution, i.e. $\pi^s(\hat{Y}_{t-1}) \propto p(\hat{y}_t | \hat{Y}_{t-1})$.

The baseline policy is used to generate a sequence $\hat{Y}^b$, whereas the sampling policy is used to generate another sequence $\hat{Y}^s$. The reward $R(\hat{Y}^s)$ is then defined as the difference between the sequence-level accuracy ($A$) of the sampled sequence $\hat{Y}^s$ and the baseline sequence $\hat{Y}^b$. The corresponding self-critical policy gradient loss is:
\begin{align}
L_{RL} & = - R(\hat{Y}^s) \times L_{MLE}(\hat{Y}^s) \nonumber \\
 & = -\! \left(\!A(\hat{Y}^s)\! -\! A(\hat{Y}^b)\!\right) \times L_{MLE}(\hat{Y}^s)
\label{eq:rl-loss}
\end{align}
Thus, minimizing the RL loss is equivalent to maximizing the likelihood of the sampled $\hat{Y}^s$ if it obtains a higher sequence-level accuracy than the baseline $\hat{Y}^b$.

\section{Experimental Evaluation}
\label{section:exp}

All models are implemented in Tensorflow using dropout to deal with overfitting. For both datasets, 10\% of the data is put aside for validation. After tuning on the artificial validation data, the feed-forward neural networks dropout rate was set to 0.5 and the LSTM units dropout rate was set to 0.3. The word embeddings had dimensionality of 64 and were initialized at random. Optimization is performed with the Adam algorithm. For each dataset, we use five-fold cross evaluation, where the data is partitioned into five folds, one fold is used for testing and the other folds for training. The process is repeated five times to obtain test results on all folds. We use an early-stop strategy on the validation set. The number of gradient updates is typically more than 20,000. All the experiments are performed on a single NVIDIA GTX1080 GPU.

The models are trained and evaluated on the artificial interactions first. To evaluate on real interactions, the models are pre-trained on the entire artificial dataset and then fine-tuned using real interactions. \textit{SPAAC-RL} is pre-trained with MLE loss to provide more efficient policy exploration. We use sequence level accuracy as evaluation metric for all models: a generated sequence is considered correct if and only if all the generated tokens match the ground truth tokens.

We report experimental evaluations of the proposed models \textit{SPAAC-MLE} and \textit{SPAAC-RL} and baseline models \textit{SeqGen}, \textit{SeqGen+Att2In} on the Real and Artificial Interactions Datasets in Table~\ref{tab:res1}. We also report examples generated by the SPAAC models in Tables~\ref{tab:example2} and \ref{tab:example1}. 

\begin{table}[t]
\centering
\renewcommand{\arraystretch}{1.1}
\begin{center}
\begin{tabular}{lcc}
\toprule[1.2pt]
{\bf Models} & Artificial & Real \\
\midrule
\textit{SeqGen} & 51.8 & 22.2 \\ 
\textit{SeqGen+Att2In} & 72.7 & 35.4 \\ 
\textit{SPAAC-MLE} & {\bf 84.3} & {\bf 66.9}\\ 
\textit{SPAAC-RL} & {\bf 88.7} & {\bf 74.8} \\
\bottomrule[1.2pt]
\end{tabular}
\caption{Sequence-level accuracy on the 2 datasets.}
\label{tab:res1}
\end{center}
\end{table}

\begin{table}[t]
\small
\begin{center}
\begin{tabular}{lccc}
\toprule[1.2pt]
Well the Finger Stick is 56.\\
T\&MLE\&RL:\\
$e.type==Fingerstick \wedge e.value==56$ \\

It looks like she suspended her pump.\\
T\&MLE\&RL:\\
$Suspended(e) \wedge around( e.time, e(-1).time)$\\

\toprule[1.2pt]
Let's look at the next day.\\
T\&MLE\&RL: $DoSetDate( currentdate+1)$\\

\toprule[1.2pt]
See if he went low.\\
T\&MLE\&RL: $Answer( any( e, hypo( e)))$\\

\toprule[1.2pt]
\toprule[1.2pt]

Let's see what kind of exercise that is, \\where the steps are high?\\
T\&RL:$Answer(e.kind)\wedge e.type==exercise$\\
$\wedge around(e.time,e_1.time)\wedge e_1.type==\boldsymbol{stepcount}$\\
$\wedge\boldsymbol{high(e_1.value)}$\\
MLE:
$Answer(e.kind)\wedge e.type==exercise$\\
$\wedge around(e.time,e_1.time)\wedge e_1.type==\boldsymbol{exercise}$\\
$\wedge\boldsymbol{e_1.type==exercise}$\\

\toprule[1.2pt]

Click on the exercise.\\
T\&RL:$\boldsymbol{DoClick}(e)\wedge e.type==exercise$\\
MLE:
$\boldsymbol{Answer}(e) \wedge e.type==exercise$\\







\bottomrule[1.2pt]
\end{tabular}
\caption{Examples generated by \textit{SPAAC-MLE} and \textit{SPAAC-RL} using real interactions. T: true logical forms. MLE: logical forms by \textit{SPAAC-MLE}. RL: logical forms by \textit{SPAAC-RL}.}
\label{tab:example2}
\end{center}
\end{table}

\begin{table}[t]
\small

\begin{center}
\begin{tabular}{lcc}

\toprule[1.2pt]
Does he always get some sleep around 4:30pm? \\
T\&MLE\&RL: $Answer( cond( around( x, 4:30pm) $\\
\hspace{0.5em}$=>any(e.type==reportedsleep\wedge e.time == x)))$\\

\toprule[1.2pt]
Is it the first week of the patient?\\
T\&MLE\&RL: $Answer( week( currentdate) == x)$\\
\hspace{0.5em} $\wedge order( x, 1, sequence( e, e.type == week))$\\

\toprule[1.2pt]
Does she ever get some rest around 5:37pm?\\
T\&MLE\&RL: $Answer( any( e.type == reportedsleep $\\
\hspace{0.5em} $\wedge around( e.time, 5:37pm)))$\\

\toprule[1.2pt]
When is the first time he changes his infusion set? \\
T\&MLE\&RL: $Answer( e.date)$\\
\hspace{0.5em} $\wedge order( e, 1, sequence( e, e.type == infusionset))$\\

\toprule[1.2pt]
\toprule[1.2pt]

How many months she has multiple exercises?\\
T\&RL:
$Answer(count(x, count(e, e.type==exercise$\\
$\wedge e.date==x)>1\wedge x.type==\boldsymbol{month}))$ \\
MLE:
$Answer(count(x, count(e, e.type==exercise$\\
$\wedge e.date==x)>1\wedge x.type==\boldsymbol{week}))$ \\

\toprule[1.2pt]

Toggle so we can see fingersticks.\\
T\&RL:
$DoToggle(on, \boldsymbol{fingersticks})$\\
MLE:
$DoToggle(on, \boldsymbol{bgl})$\\




\bottomrule[1.2pt]
\end{tabular}
\caption{Examples generated by \textit{SPAAC-MLE} and \textit{SPAAC-RL} using artificial interactions. T: true logical forms. MLE: logical forms generated by \textit{SPAAC-MLE}. RL: logical forms generated by \textit{SPAAC-RL}.}
\label{tab:example1}
\end{center}
\end{table}

\subsection{Discussion}
\label{section:dis}


The results in Table~\ref{tab:res1} demonstrate the importance of modeling context-dependency, as the two {\it SPAAC} models outperform the baselines on both datasets. The RL model also obtains substantially better accuracy than the MLE model. 
The improvement in performance over the MLE model for the real data is statistically significant  at $p = 0.05$ in a one-tailed paired t-test. 

Analysis of the generated logical forms revealed that one common error made by \textit{SPAAC-MLE} is the generation of incorrect event types. Some of these errors are fixed by the current RL model. 
However, there are instances where even the RL-trained model outputs the wrong event type. By comparing the sampled logical forms $\hat{Y}^{s}$ and the generated baseline logical forms $\hat{Y}^{b}$, we found that sometimes the sampled tokens for event types are the same as those in the baseline. An approach that we plan to investigate in future work is to utilize more advanced sampling methods to generate $\hat{Y}^{s}$, 
in order to achieve a better balance between exploration and exploitation.

\section{Related Work}
\label{section:related}


Question Answering has been the topic of recent research \cite{yih2014semantic, dong2015question,andreas2016learning,hao2017end,abujabal2017quint,chen2017exploration}.
Semantic parsing, which maps text in natural language to meaning representations in formal logic, has emerged as an important component for building QA systems,
as in ~\cite{liang2016learning,jia2016data,zhong2017seq2sql}. 
Context-dependent processing has been explored in complex, interactive QA \cite{harabagiu2005experiments,kelly2007overview} and semantic parsing \cite{zettlemoyer2009learning,artzi2011bootstrapping,iyyer2017search,suhr2018learning,long2016simpler}. Although these approaches take into account sequential dependencies between questions or sentences, the setting in our work has a number of significant distinguishing features, such as the importance of time -- data is represented naturally as multiple time series of events -- and the anchoring on a graphical user interface that also enables direct interactions through mouse clicks and a combination of factual queries and interface commands.

\citet{dong2016language} use an attention-enhanced encoder-decoder architecture to learn the logical forms from natural language without using hand-engineered features. Their proposed Seq2Tree architecture can capture the hierarchical structure of logical forms. \citet{jia2016data} train a sequence-to-sequence RNN model with a novel attention-based copying mechanism to learn the logical forms from questions. The copying mechanism has been investigated by~\citet{gu2016incorporating} and~\citet{gulcehre2016pointing} in the context of a wide range of NLP applications. These semantic parsing models considered sentences in isolation. In contrast, generating correct logical forms in our task required modeling sequential dependencies between logical forms. In particular, coreference is modeled between events mentioned in different logical forms by repurposing the copying mechanism originally used for modeling out-of-vocabulary tokens.

\section{Conclusion}

We introduced a new semantic parsing setting in which users can query a system using both natural language and direct interactions (mouse clicks) within a graphical user interface. Correspondingly, we created a dataset of real interactions and a much larger dataset of artificial interactions. The correct interpretation of a natural language query often requires knowledge of previous interactions with the system. We proposed a new sequence generation architecture that modeled this context dependency through multiple attention models and a copying mechanism for solving coreference. The proposed architecture is shown to outperform standard LSTM encoder-decoder architectures that are context agnostic. Furthermore, casting the sequence generation process in the framework of reinforcement learning alleviates the exposure bias and leads to substantial improvements in sequence-level accuracy.

The two datasets and the implementation of the systems presented in this paper are made publicly available at \url{https://github.com/charleschen1015/SemanticParsing}. The data visualization GUI is available under the name {\sc OhioT1DMViewer} at \url{http://smarthealth.cs.ohio.edu/nih.html}.

\section*{Acknowledgments}

This work was partly supported by grant 1R21EB022356 from the National Institutes of Health. We would like to thank Frank Schwartz and Cindy Marling for contributing real interactions, Quintin Fettes and Yi Yu for their help with recording and pre-processing the interactions, and Sadegh Mirshekarian for the design of the artificial data generation. We would also like to thank the anonymous reviewers for their useful comments.


\bibliography{cameraready}
\bibliographystyle{acl_natbib}

\end{document}



\appendix

\section{Artificial Interactions}
\label{sec:appendix_artificial}

An artificial data generator was designed and implemented to simulate doctor-system interactions, with sentence \textit{templates} defining the skeleton of each entry in order to maintain high-quality sentence structure and grammar. We used a context free grammar to implement a \textit{template language} that can specify a virtually unlimited number of templates and generate as many examples as desired. Below we show a simplification of three sample rules from the grammar:
\begin{align}
\langle S \rangle &\rightarrow \mbox{maximum heart rate on } \langle P \rangle \mbox{ today}? \nonumber\\ 
\langle P \rangle &\rightarrow \mbox{the day before } \langle P \rangle \nonumber\\
\langle P \rangle &\rightarrow \mbox{the Monday after} \nonumber
\end{align}
A sample derivation using these rules is "maximum heart rate on the Monday after today?". 

The implementation allows for the definition of any number of non-terminals, which we call \textit{types}, and any number of {\it templates}, which are the possible right hand sides of the starting symbol $S$. The doctor-system interactions can be categorized into three types: {\it questions}, {\it statements}, and {\it clicks}, where templates can be defined for each type, as shown in Table~\ref{tab:template_examples}. Given the set of types and templates, a virtually unlimited number of sentences can be derived to form the artificial dataset. Since the sentence generator chooses each template randomly, the order of sentences in the output dataset will be random. However, two important aspects of the real interactions are context dependency and coreference. To achieve context dependency, the implementation allows for more complex \textit{combo templates} where multiple templates are forced to come in a predefined order. It is also possible to specify groups of templates, via tagging, and combine groups rather than individual templates. Furthermore, each NL sentence template is paired with a LF template, and the two templates are instantiated jointly, using a reference mechanism to condition the logical form generation on decisions made while deriving the sentence.

\begin{table}[t]
\setlength{\tabcolsep}{0em}
\small
\centering
\begin{tabular}{l}
	\hline
	\multicolumn{1}{c}{Example types.} \\
	\hline
	week\_days$\rightarrow$Monday\,|\,Tuesday\,|\,...\,|\,Sunday\\
	daily\_intervals$\rightarrow$morning\,|\,afternoon\,|\,evening\,|\,night\\
	daily\_intervals\_logic$\rightarrow$ Morning\,|\,Afternoon\,|\,Evening\,|\,Night\\
	any\_event$\rightarrow$heart rate\,|\,bolus\,|\,blood glucose level\\
	any\_event\_logic$\rightarrow$HeartRate\,|\,Bolus\,|\,BGL\\
	\hline
	\hline
	\multicolumn{1}{l}{\textbf{Example 1}: a statement, involving referencing} \\
	\hline
	Let's go to [week\_days]. $\rightarrow DoSetDate([\$1])$\\ 
	Possible derivations: \\
	\hspace{1em} $\bullet$ Let's go to Monday. $\rightarrow DoSetDate(\mbox{Monday})$ \\
	\hspace{1em} $\bullet$ Let's go to Tuesday. $\rightarrow DoSetDate(\mbox{Tuesday})$\\[0.5em]
	\hline
	
	\multicolumn{1}{l}{\textbf{Example 2}: a combo statement capturing temporal dependence} \\
	\hline
	[[let's/please/we can]/can we] turn the [any\_event] off[\$1:./?]\\
	\hspace{1em} $DoToggle(\mbox{Off}, [\$2:\mbox{any\_event\_logic}])$\\
	\ldots and the [any\_event] too. \\
	\hspace{1em} $DoToggle(\mbox{Off}, [\$1:\mbox{any\_event\_logic}])$ \\
	Possible derivations: \\
	\hspace{1em} $\bullet$ please turn the bolus off. $\rightarrow DoToggle(\mbox{Off}, \mbox{Bolus})$ \\
	\hspace{2em} and the heart rate too. $\rightarrow DoToggle(\mbox{Off}, \mbox{HeartRate})$ \\
	\hspace{1em} $\bullet$ can we turn the blood glucose level off?\\
	\hspace{11em} $\rightarrow DoToggle(\mbox{Off}, \mbox{BGL})$\\
	\hspace{2em} and the bolus too.$\rightarrow DoToggle(\mbox{Off}, \mbox{Bolus})$ \\[0.5em]
	\hline
	
	\multicolumn{1}{l}{\textbf{Example 3}: a click, involving the special type clocktime} \\
	\hline
	$Click(e) \wedge e.type==[\mbox{any\_event\_logic}]$\\
	\hspace{3.8em} $ \wedge \: e.time==[\mbox{clocktime}]$\\ 
	A possible derivation: \\
	\hspace{1em} $\bullet$ $Click(e) \wedge e.type==\mbox{Bolus} \wedge e.time==\mbox{12:36 PM}$ \\[0.5em]
	\hline
	
	\multicolumn{1}{l}{\textbf{Example 4}: a question, involving the special type range} \\
	\hline
	is there [a/any][valued\_event]\\ \hspace{8em}[more/less] than [range(-500,500)]?\\
	$Answer(Any(d.value[\$3\!:>\!\!/\!\!<][\$4]$\\
	\hspace{8em}$\wedge d.type\!\!==\!\![\$2\!\!:\!\!\mbox{valued\_event\_logic}]))$\\
	One possible derivation: \\
	\hspace{1em} $\bullet$ is there any heart rate less than 250?\\ 
	\hspace{1em} $Answer(Any(d.value<250 \wedge d.type==\mbox{HeartRate}))$ \\[0.5em]
	\hline
\end{tabular}
\caption{Examples of generation of artificial samples.}
\label{tab:template_examples}
\vspace{-6mm}
\end{table}

Table~\ref{tab:template_examples} shows examples of how artificial sentences and their logical forms are generated given templates and types. Most types are defined using context free rules. There are however special types, such as [clocktime] and [range()], which are dynamically rewritten as a random time and a random integer from a given range, as shown in Examples 3 and 4, respectively. Note that most examples use referencing, which is a mechanism to allow for dynamic matching of terminals between the NL and LF derivations. In Example 1, $\$1$ in the logical form template refers to the first type in the main sentence, which is [week\_days]. This means that whatever value is substituted for [week\_days] should appear verbatim in place of $\$1$. In case a coordinated matching from a separate list of possible options is required, such as in Example 2, another type can be selected. In Example 2, [\$2:any\_event\_logic] will be option $i$ from the type [any\_event\_logic] when option $i$ is chosen in the main sentence for the second template, which is [any\_event]. 
